\documentclass[11pt]{style/prism-princeton}

\makeatletter
\newcommand{\longdash}[1][2em]{%
  \makebox[#1]{$\m@th\smash-\mkern-7mu\cleaders\hbox{$\mkern-2mu\smash-\mkern-2mu$}\hfill\mkern-7mu\smash-$}}
\makeatother
\newcommand{\omitskip}{\kern-\arraycolsep}

\author{Mingtong Zhang}
\author{Dhruv Shah}

\affiliation{Princeton University}

\usepackage{algorithm}
\usepackage{algpseudocode}
\usepackage{wrapfig}

\newcommand{\MethodName}{VERITAS~}
\begin{document}

\title{Visual Verification Enables Inference-time Steering and Autonomous Policy Improvement}

\abstract{
Robots deployed in the real world should learn from their experience and improve over time. This requires a mechanism of practicing and learning from feedback. In this paper, we propose VERITAS, a generator-verifier framework for generalist robot policies for inference-time policy steering and self-improvement. We use a pre-trained generalist robot policy as a ``generator'' and pair it with a gradient-free ``visual verifier'' that evaluates actions at inference time. This framework enables inference-time steering that improves policy performance without additional training. We demonstrate that inference-time verification consistently outperforms vanilla generalists without training on additional demonstration data. Additionally, we demonstrate that the verified rollouts provide effective supervision for offline policy improvement: policies fine-tuned on verified self-generated trajectories achieve consistent performance gains. Notably, we find that post-training with verified rollouts achieves comparable efficiency to expert demonstrations, while requiring no human interventions. Our results highlight inference-time verification as a practical and scalable mechanism for improving robotic policies during deployment.
}


\website{https://veritas-improvement.github.io/}{veritas-improvement.github.io}

\code{https://github.com/prism-princeton/project}{github.com/prism-princeton/project}

\maketitle


\section{Introduction}
\label{sec:intro}

\begin{figure*}[t]
    \centering
    \includegraphics[width=\linewidth]{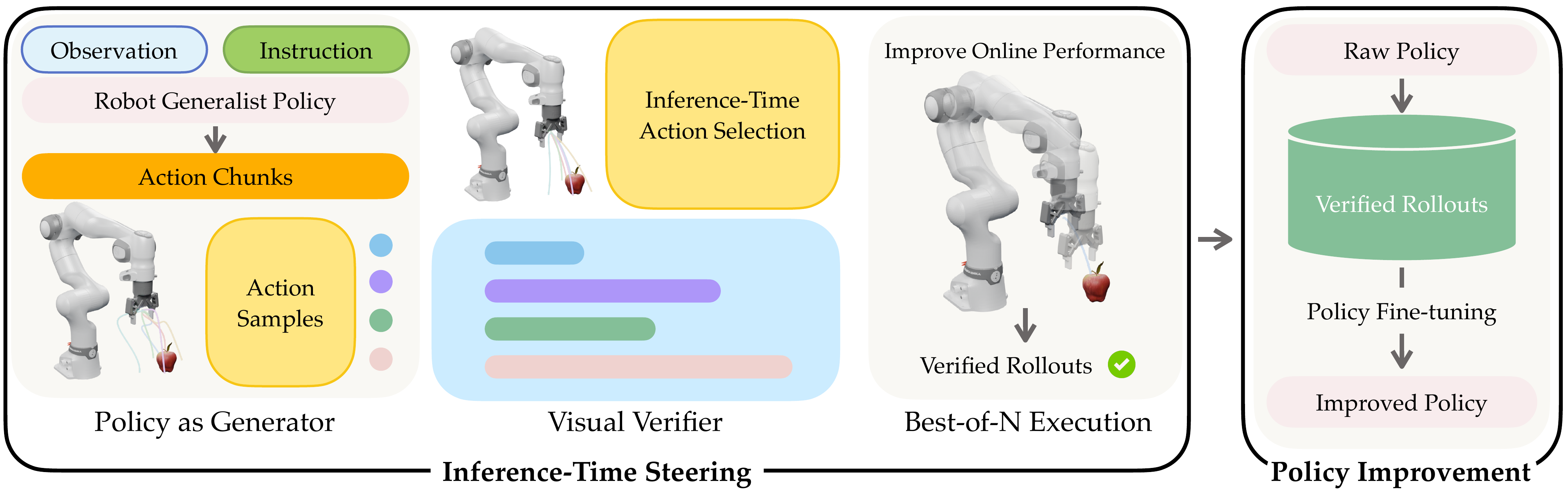}
    \caption{\textbf{Overview of VERITAS.} A pre-trained generalist policy acts as a stochastic generator, sampling multiple short-horizon action chunks at each decision step. A gradient-free visual verifier scores these candidates based on task alignment and physical plausibility, and the highest-scoring action is executed, yielding immediate performance gains at inference time. Successful verifier-guided rollouts are logged and reused for offline policy improvement, forming a data flywheel that distills verification-time reasoning into the policy and enables continual improvement with minimal human supervision.}
    \label{fig: teaser}
\end{figure*}

Robotic foundation models trained on large-scale demonstration data~\cite{team2025gemini, o2024open,  barreiros2025careful, bjorck2025gr00t, blackpi0, zitkovich2023rt, molmospaces2026} have shown advanced capabilities in various domains, such as manipulation and navigation~\cite{deshpande2026molmobot, liu2025rdt, kim24openvla, hirose2025omnivla, shah2022gnm}. These advancements are largely powered by large-scale robotic data collection from different sources. However, unlike texts or images, robotic data collection remains expensive, slow, and limited by the availability of human demonstrators. Thus, robot foundation models face a fundamental bottleneck: state-of-the-art models are primarily trained on data collected by human experts, which scales linearly in human expert time. This raises a key challenge: how can we improve robotic policies without scaling up the amount of human data? Our key insight is to enable robotic systems to explore the real world and learn from their experience, with minimal supervision and cost.

A promising solution inspired by recent advances in large language modeling is to scale \emph{inference-time computation} for verification~\cite{snell2025scaling, chen2024are, li2025s}. Text-based reasoning models can generate multiple solution traces for a problem and select the best one using a ``verifier''~\cite{cobbe2021training, lightman2023let, yao2023tree}. Analogous to this, perhaps a robot policy can generate and verify multiple action candidates at inference-time before committing to a physical action. Through selecting the \emph{best-of-$N$} samples using a verifier or value function, a pre-trained model might achieve performance improvements without any additional training. This insight is particularly valuable in robotic applications where data acquisition is significantly more expensive than computation. Rather than collecting additional data to improve coverage or recovery behaviors~\cite{kelly2019hg, hu2025rac}, we can instead use inference-time sampling to enable the policy to ``try'' multiple times and only execute the best generation. This allows us to trade compute for improved performance through inference-time steering. As the robot experiences successful trials, this autonomously generated data can be used to guide further improvements.

In this work, we introduce Visual \textbf{Ver}ification for \textbf{I}nference-\textbf{t}ime \textbf{S}teering and \textbf{A}utonomous Policy Improvement, or \textbf{VERITAS}, 
a robotic self-improvement method built around a \emph{generator-verifier framework}. We treat robot policies as a \emph{generator} capable of sampling diverse candidate actions, and pair it with a \emph{verifier} based on a Vision-Language-Model (VLM) which evaluates these candidates to guide exploration and task execution. This framework enables robots to autonomously explore the physical world with inference-time verification, steering existing generalist policies in an \emph{online} manner out-of-the-box. Additionally, the verified self-generated rollouts serve as high-quality training data for \emph{offline} policy improvement. Notably, this generator–verifier framework enables policy self-improvement without (i) training any additional embodied reasoning models, and (ii) requiring expensive human annotation or data collection. Instead, it externalizes reasoning from the policy itself by introducing an explicit verification stage that evaluates candidate actions at inference time. By creating a flywheel where the robot learns from its own successful executions, we ensure that the post-training data is \emph{on-policy} and physically feasible; we posit that this enables the policy to rapidly improve its behaviors and learn new behaviors, guided by the verifier. Our experiments demonstrate that our framework can improve policy performance while matching the efficiency of learning from human expert demonstrations.

We evaluate the efficacy of VERITAS for inference-time steering and autonomous policy improvement extensively in simulation~\cite{li24simpler} and in the real world~\cite{khazatsky2024droid}. Across various base policies and manipulation tasks, we find that inference-time steering with VERITAS consistently improves performance over the base policy, without modifying policy parameters. Additionally, we show that verifier-generated rollouts form an effective supervision signal for policy improvement, achieving an average improvement of $10\%$ over the base policy in simulation. 
We further find that performance gains from ``self-improvement'' (post-training on policy-generated data) can match the gains from post-training on human-expert data in the real world, suggesting that verifier-curated data is comparable to human data for autonomous policy improvement.

\section{Related Works}

Prior work has extensively studied data collection through \emph{human-in-the-loop} learning~\cite{luo2025precise, liu2022robot, zhang2019leveraging} and \emph{shared autonomy}~\cite{Dragan-2013-7699, javdani2015shared, reddy2018shared}, where robots improve by incorporating expert interventions, corrections, or demonstrations during deployment. Classic approaches such as DAgger rely on expert relabeling to correct policy mistakes on-policy~\cite{ross2011reduction, kelly2019hg, xu2025compliant}, while shared autonomy frameworks blend human control with autonomous execution to improve task success and collect corrective data. Recent systems refine this paradigm by structuring how corrective data is collected in long-horizon tasks; for example, Hu et al.~\cite{hu2025rac} explicitly decompose robot behavior into nominal execution and recovery phases, triggering human intervention when the policy enters failure states to improve downstream learning~\cite{barreiros2025careful}. Instead of using human intervention~\cite{Spencer2020LearningFI, kelly2019hg} as the source of corrective signal, we leverage \emph{execution-time verification} to autonomously evaluate candidate actions and select successful behaviors. The verifier replaces the role of the human supervisor by providing a task-aligned preference signal at inference time, allowing the robot to improve performance online and collect success-conditioned trajectories without shared control.

Inspired by advances in language modeling, where scaling test-time compute via verification and search often outperforms scaling model parameters~\cite{snell2025scaling, yao2023tree, cobbe2021training}, the robotics community has begun exploring mechanisms for embodied control. Recent work integrates explicit reasoning into robot policies by training VLAs with supervision over intermediate visual and semantic representations, enabling multi-step, visually grounded decision making~\cite{zawalski2024robotic, zhao2025cot, lee2025molmoact}.


While these works demonstrate impressive embodied reasoning capabilities, these methods rely on expensive annotations during policy pre-training and are not suitable for inference-time steering or improvement of existing policies. Another promising set of directions is sampling and steering online samples, such as Kwok et al.~\cite{kwok2025robomonkey} demonstrating that action error scales with the number of actions sampled at inference time using a learned verifier, Kwok et al.~\cite{kwok2026scaling} suggesting scaling verification leads to more effective alignment for policies, Wu et al.~\cite{wu2025you} introducing a runtime steering method that verifies actions against a reasoning VLA's self-generated textual plan, and Nakamoto et al.~\cite{nakamoto2024steering}, which trains a language-conditioned value function using offline RL and uses it for steering the generalist policy. ~\cite{wu2025foresight, yuan2026actasklearnuncertaintyaware} requires world models' prediction to select reasonable actions for policy steering. While these methods demonstrate the power of test-time sampling, they typically treat inference-time search as a ``momentary boost'', relying on expensive verification, that does not necessarily improve the policy in long-term deployment. Our work takes a novel perspective by treating online verification as a data engine without training task-specific verifiers, and allowing us to distill expensive inference-time compute into the base policy to enhance its capabilities via autonomous verification and post-training. Our experiments further show that our visual verification framework outperforms various other test-time sampling approaches for robot policies.

Long-term deployment of generalist robots requires policies to continually learn and adapt from their experience. However, standard imitation learning suffers from covariate shift, where policy performance degrades as the robot drifts from the expert's state distribution~\cite{ross2011reduction}. Traditional solutions~\cite{ross2011reduction, kelly2019hg} mitigate this by querying human experts for labels on robot-visited states. However, these human-in-the-loop paradigms are difficult to scale. Our approach replaces the human labeler with an automated verifier, ensuring the training data is strictly \emph{on-policy} and physically feasible for the robot's embodiment. This effectively mitigates the distribution shift problem by aligning the training distribution with the support of the robot's exploration prior, unlocking scalable self-improvement without human intervention. 
\section{Preliminaries}
\label{sec:preliminaries}

We formulate the problem of language-guided robotic manipulation as a Partially Observable Markov Decision Process (POMDP)~\cite{Sutton1998} defined by the tuple $(\mathcal{S}, \mathcal{O}, \mathcal{A}, \mathcal{T}, \mathcal{R}, \gamma)$.

\vspace{2mm}
\noindent \textbf{Problem Formulation.} At each sampling timestep $t$, the robot receives an observation $o_t \in \mathcal{O}$ and a natural language instruction $l \in \mathcal{L}$. Our objective is to train a policy $\pi_\theta$ that maximizes the expected success rate over a distribution of tasks. Unlike standard imitation learning, which assumes a fixed static dataset, our objective is to enable the policy to self-improve its performance post-deployment by collecting and learning from its own experience.

\vspace{2mm}
\noindent \textbf{Generalist Robot Policies.} In this work, we assume access to a pre-trained ``generalist policy'' $\pi_\theta$, that can be conditioned on the current observation and a language instruction, and predicts actions $a_t$ to control a robot. We assume that $\pi_\theta$ can successfully solve a large number of tasks in the target environment, and are agnostic to the specific architecture or data sources the policy was pre-trained on. Following recent advancements in high-frequency control, we employ \emph{action chunking}~\cite{zhao2023act, chi2023diffusion}: the policy predicts a sequence of actions of horizon $H$:
\begin{equation}
    \boldsymbol{a}_{t:t+H} \doteq (a_t, a_{t+1}, \dots, a_{t+H-1}) \in \mathcal{A}^H.
\end{equation}

\vspace{2mm}
\noindent \textbf{Generative Policies as Priors.} Our method relies on the policy's ability to model complex, multi-modal action distributions (e.g., grasping an object from different angles, solving a multi-step task in different orders, etc.). We focus on generative policies trained with flow-matching or diffusion objectives~\cite{blackpi0, chi2023diffusion}. Unlike deterministic regression policies that output a single mean action, a generative policy learns the conditional probability distribution $p(\boldsymbol{a}_{t:t+H} \mid o_t, l)$. At inference time, action chunks are generated via a stochastic sampling process. In this work, we leverage this stochasticity not just for robustness, but as a mechanism for exploring diverse, functional affordances~\cite{smith2024STEER} of the scene.
\section{A Generator-Verifier Framework for Robot Policies}
\label{sec:method}

    
\begin{figure*}[t]
    \centering
    \includegraphics[width=\linewidth]{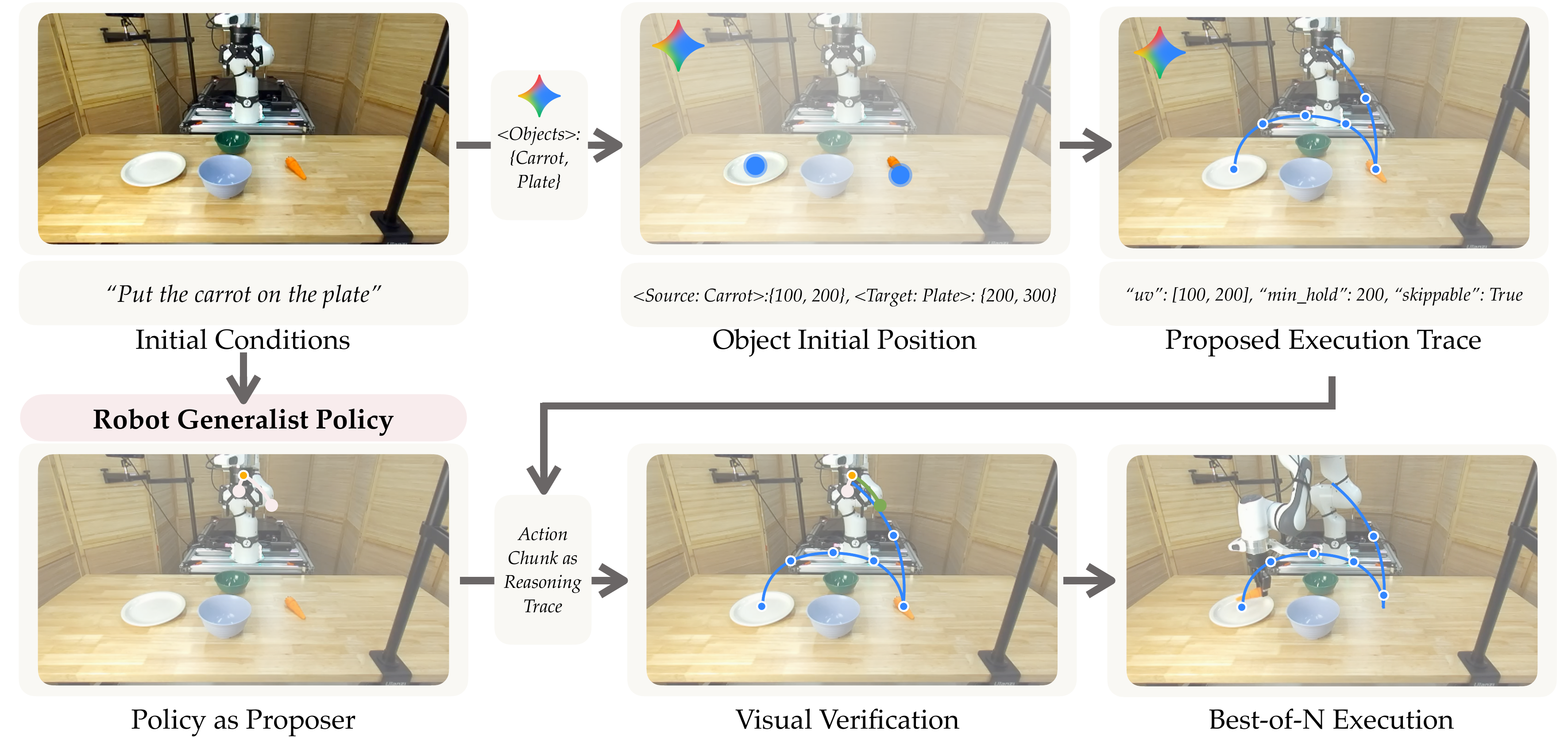}
    \caption{\textbf{The overview of \MethodName for inference-time steering.} At inference time, a two-call VLM scheme is used to construct a visual verifier. First, given the initial observation and task instruction, the VLM identifies the target object to focus on, a detection model is then used to localize the object in image space. Second, conditioned on the object location, instruction, and initial observation, the VLM proposes a sequence of pixel-space waypoints that define a full visual trace for the task. During execution, the policy samples multiple action candidates, which are scored based on how well their resulting motion follows the proposed visual trace, and the best candidate is selected for execution. Successful verifier-guided rollouts are logged and used as training data for offline policy fine-tuning, resulting in an improved policy with higher task performance.}
    \label{fig: method_overview}
    \vspace{-10pt}
\end{figure*}

We present a \emph{generator--verifier} inference framework that steers a generalist policy with execution-time selection. The core design principle is to utilize (i) \emph{generation} by a generalist policy $\pi_\theta$ as a stochastic generator, and (ii) \emph{evaluation} by a plug-in verifier $V$ that is policy-agnostic and gradient-free. At every decision step, the robot samples multiple short-horizon action chunks, scores them with $V$, and executes the best candidate. This yields immediate online performance gains without updating $\theta$, while producing success-conditioned trajectories for offline policy improvement. The complete procedure is detailed in Algorithm~\ref{alg:generator_verifier_offline}.

\subsection{The Policy as a Stochastic Generator}
\label{subsec:generator}

Standard policy deployment typically utilizes deterministic or greedy decoding to minimize variance. We instead view the pre-trained policy $\pi_\theta$ as a \textit{stochastic generator}. We posit that a generalist policy trained on diverse data maintains a broad ``prior" over valid affordances, and by sampling from this distribution, we can encourage and recover high-quality behaviors that might have lower initial probability but higher task alignment.

Formally, at discrete sampling timestep $t$, given observation $o_t$ and language instruction $l$, the policy generates a set of $N$ candidate action chunks (Algorithm~\ref{alg:generator_verifier_offline}, Line~\ref{alg:sample}):
\begin{equation}
    \boldsymbol{a}_{t:t+H}^{(i)} \sim \pi_\theta(\cdot \mid o_t, l), \quad i \in \{1, \dots, N\}
    \label{eq: samples}
\end{equation}
where each candidate $\boldsymbol{a}_{t:t+H}^{(i)} = (a_t, a_{t+1}, \dots, a_{t+H-1})$ represents a sequence of $H$ actions. Utilizing action chunks is critical; it amortizes the cost of verification over the horizon $H$ and provides the verifier with temporal context to evaluate the consequence of a behavior, rather than a micro instantaneous movement.

\subsection{The Verifier Interface}
\label{subsec:verifier}

To select the optimal candidate from the generator, we introduce a verifier $V$. We define the verifier as a plug-and-play, gradient-free function that maps the observation, candidate action chunk, and task instruction to a scalar score:
\begin{equation}
    V(o_t, \boldsymbol{a}_{t:t+H}, l) \in \mathbb{R}.
    \label{eq:verifier}
\end{equation}
Conceptually, $V$ estimates the utility or alignment of executing $\boldsymbol{a}_{t:t+H}$ from state $o_t$ under instruction $l$. Importantly, steering with $V$ operates solely at inference time and does not update $\pi_\theta$. As a result, the verifier functions as a plug-and-play module that can be instantiated with various mechanisms (e.g., VLMs, geometric constraints, or learned value models) without retraining the base policy.

\subsection{Inference-Time Steering}
\label{subsec:steering}

We combine the generator and verifier via \textit{Best-of-$N$} selection. This mechanism trades inference-time computation for performance, allowing the robot to ``think" (sample and verify) before acting.

As outlined in Algorithm~\ref{alg:generator_verifier_offline}, the control loop proceeds as follows:
\begin{enumerate}
    \item \textbf{Sample:} The policy generates $N$ diverse action chunks $\{\boldsymbol{a}_{t:t+H}^{(i)}\}_{i=1}^N$.
    \item \textbf{Verify:} The verifier scores each candidate: $v_i = V(o_t, \boldsymbol{a}_{t:t+H}^{(i)}, l)$.
    \item \textbf{Select:} The highest-scoring candidate index is selected:
    \begin{equation}
        i^\star = \arg\max_{i \in \{1,\dots,N\}} v_i, \quad \boldsymbol{a}^\star_{t:t+H} = \boldsymbol{a}^{(i^\star)}_{t:t+H}
        \label{eq:select-candidate}
    \end{equation}
    \item \textbf{Execute:} The robot executes the chunk $\boldsymbol{a}_{t:t+H}^\star$ in the environment.
\end{enumerate}

This steering process serves as a filter, prioritizing physically valid and task-aligned actions. Importantly, because $\pi_\theta$ is fixed, these performance gains come purely from inference-time computation.

\subsection{Autonomous Self-Improvement}
\label{subsec:self_improvement}

While inference-time steering improves performance, it incurs a computational cost at every decision step. To amortize this cost and permanently improve the policy, we utilize the verified trajectories as a source of autonomous supervision with no extract human expert efforts.

This process creates a \emph{data flywheel}. During policy deployment, we log the successful verified trajectories generated via inference-time steering into a dataset $\mathcal{D}_{\text{auto}}$.
\begin{equation}
    \mathcal{D}_{\text{auto}} = \bigcup_{k} \{(o_t, \boldsymbol{a}_{t:t+H}^\star, l)\}_{t \in \mathcal{I}_{\text{success}}}
    \label{eq:grow-dataset}
\end{equation}
We then fine-tune the original policy $\pi_\theta$ on $\mathcal{D}_{\text{auto}}$ using standard supervised behavior cloning to obtain updated parameters $\theta'$:
\begin{equation}
    \theta' \leftarrow \arg\min_\theta \mathbb{E}_{\mathcal{D}_{\text{auto}}} [-\log \pi_\theta(\boldsymbol{a}_{t:t+H}^\star \mid o_t, l)]
    \label{eq:post-training}
\end{equation}

\vspace{2mm}
\textbf{Mitigating Distribution Shift.} A fundamental challenge in imitation learning is the distribution shift between the expert and the learner. Standard human-in-the-loop approaches require expensive expert relabeling. Our framework automates this: because the data in $\mathcal{D}_{\text{auto}}$ was generated by the robot's own policy (the generator), it is inherently on-policy and kinematically feasible. The verifier acts as the "expert" filter, ensuring only high-quality data is retained. This closes the loop: the reasoning performed by the verifier is distilled into the policy, reducing the need for large $N$ samples in future deployments. Importantly, as the verified data is generated by policy itself, the action distribution is close to the policy's action prior, which mitigates the shift and remains efficient for policy improvement.

\begin{algorithm}[t]
\caption{\MethodName Online Steering and Autonomous Policy Improvement}
\label{alg:generator_verifier_offline}
\SetKwInOut{Input}{Input}
\SetKwInOut{Output}{Output}

\Input{Policy $\pi_\theta$, verifier $V$, instruction $l$, horizon $H$, samples $N$}
\Output{Executed trajectory, dataset $\mathcal{D}_{\text{auto}}$, updated policy $\pi_{\theta'}$}

\BlankLine
\tcp{Phase 1: Inference-time Steering}
Initialize $\mathcal{D}_{\text{auto}} \leftarrow \emptyset$ and observe initial state $o_0$\;

\For{$t = 0, H, 2H, \dots$ until termination}{
    \tcp{1. Sample diverse candidates (Eq~\ref{eq: samples})}
    Sample $\{\boldsymbol{a}^{(i)}_{t:t+H}\}_{i=1}^N$ where $\boldsymbol{a}^{(i)}_{t:t+H} \sim \pi_\theta(\cdot \mid o_t,l)$\; \label{alg:sample}
    
    \tcp{2. Verify candidates}
    Score each candidate $v_i \leftarrow V(o_t,\boldsymbol{a}^{(i)}_{t:t+H},l)$\; \label{alg:score}
    
    \tcp{3. Select Best-of-N (Eq.~\ref{eq:select-candidate})}
    Select $i^\star \leftarrow \arg\max_i v_i$ and set $\boldsymbol{a}^\star_{t:t+H} \leftarrow \boldsymbol{a}^{(i^\star)}_{t:t+H}$\; \label{alg:select}
    
    \tcp{4. Execute and Log (Eq.~\ref{eq:grow-dataset})}
    Execute $\boldsymbol{a}^\star_{t:t+H}$ and observe $o_{t+H}$\; \label{alg:execute}
    Log $(o_t,\boldsymbol{a}^\star_{t:t+H},l)$ into $\mathcal{D}_{\text{auto}}$\; \label{alg:log}
}

\BlankLine
\tcp{Phase 2: Offline Self-Improvement}
\If{Offline update is enabled}{ \label{alg:offline_start}
    Initialize $\theta' \leftarrow \theta$\;
    \For{$e = 1$ \KwTo $E$}{
        Sample batch $\mathcal{B} \sim \mathcal{D}_{\text{auto}}$\;
        Update $\theta'$ via Behavior Cloning (Eq.~\ref{eq:post-training}):
        \begin{equation*}
        \theta' \leftarrow \theta' - \eta \nabla_{\theta'} \sum_{(o,\boldsymbol{a},l)\in \mathcal{B}} \mathcal{L}_{\text{BC}}\!\left(\pi_{\theta'}(\cdot\mid o,l), \boldsymbol{a}\right)
        \end{equation*}
    }
    \Return $\pi_{\theta'}$\; \label{alg:offline_end}
}
\end{algorithm}
\section{Instantiating \MethodName for Robotic Manipulation}
\label{sec:instantiation}

We now describe a concrete instantiation of the proposed generator-verifier framework for tabletop robotic manipulation. In this work, as an example, we conduct our analysis using the $\pi_0$ architecture~\cite{blackpi0}, an open-source Vision-Language-Action (VLA) model trained on thousands of hours of diverse robot trajectories. $\pi_0$ utilizes a mixture-of-transformers backbone with a flow-matching action head. Crucially, this flow-matching formulation allows us to effectively sample complex, multi-modal action distributions, providing the necessary diversity to explore valid solutions at test time. We pair this generator with our visual verifier designed to ground high-level semantic instructions into precise geometric constraints.

\subsection{Visual Verifier Design}
\label{subsec:visual_verifier}

To ground high-level semantic instructions into precise geometric constraints without requiring expensive 3D annotations, we design a verifier based on absolute pixel-space trajectories proposed by a frontier vision-language model (VLM)~\cite{comanici2025gemini}. This design leverages the strong spatial and embodied reasoning capabilities of modern VLMs to create a ``visual guardrail'' for the policy.

The verification process consists of two stages:
\begin{enumerate}
    \item \textbf{Trace Generation:} At the onset of a task, the VLM is provided with the initial observation $o_0$ and the language instruction $l$. It is prompted to propose a \emph{visual reasoning trace}: a sequence of pixel-space waypoints $\{w_k\}$ overlaid on the image that defines the ideal end-effector trajectory for task completion. This trace remains static throughout execution to ensure temporal consistency.
    
    \item \textbf{Geometric Scoring:} At inference time, for each candidate action chunk $\boldsymbol{a}_{t:t+H}^{(i)}$ sampled from $\pi_\theta$, we project the resulting end-effector/gripper positions into the image plane. The verifier score $v_i$ is computed as the negative Euclidean distance between the candidate's projected trajectory and the nearest segment of the VLM-generated trace.
\end{enumerate}

Crucially, this design allows for efficient verification without requiring online tracking or expensive VLM querying at every verification step. Once the visual trace is generated, verification reduces to a fast geometric consistency check, enabling the evaluation of large $N$ samples with minimal latency overhead. We also have other types of verifier design and ablate them in the experiments. For more details on our verifier and alternate designs, please see Appendix~\ref{app:verifier}.

\subsection{Steering at Inference-Time}
\label{subsec:inference_steerability}

We implement a simple inference-time steering mechanism using the Best-of-$N$ strategy described in Section~\ref{subsec:steering}. 

\vspace{2mm}
\textbf{Implementation Details.} As an example, we deploy the $\pi_0$-DROID policy on a workstation with one NVIDIA GeForce RTX 5090 GPU. At each control frequency step (e.g., 15 Hz), we define an action chunk horizon of $H$ steps. The generator samples $N$ candidates in parallel. The visual verifier evaluates these $N$ chunks against the pre-computed visual trace.

\vspace{2mm}
\textbf{Latency and Throughput.} The VLM verifier generates an execution trace only once at the start of episode, and hence, is not run in-the-loop with the policy. The geometric verification incurs negligible overhead ($<1$ ms) and runs online. The cost of policy inference scales linearly with the number of sampled actions $N$, and can be batched during inference. For our experiments, we use $N=5$ samples and maintain a control frequency of 15 Hz. This trade-off allows us to filter out physically invalid or erratic behaviors that the base policy might occasionally produce.


\subsection{Policy Improvement}
\label{subsec:policy_improvement}

To close the loop, we utilize the successful verified trajectories curated by our steerable system for offline policy fine-tuning.

\textbf{Data Collection.} We deploy the steered policy in the real world to solve a diverse set of manipulation tasks (Section~\ref{sec:experiments}). We log all executed action chunks $\boldsymbol{a}_{t:t+H}^\star$ that lead to successful task completion after verification, filtering out failures. This results in a high-quality dataset $\mathcal{D}_{\text{auto}}$.

\textbf{Training Setup.} We fine-tune the pre-trained $\pi_0$-DROID checkpoint on $\mathcal{D}_{\text{auto}}$ using the behavior cloning objective (Eq.~\ref{eq:post-training}). We use a learning rate of $5\times 10^{-5}$ and train for $20,000$ steps. As shown in our experiments, even small amounts of this verifier-curated data (e.g., 20 trajectories) are sufficient to significantly shift the policy distribution towards the robust behaviors discovered during steering, effectively distilling the VLM's reasoning into the policy's weights.
\section{Experiments}
\label{sec:experiments}

In our experiments, we aim to assess the efficacy of the generator-verifier paradigm as a data flywheel for policy improvement. We study the following research questions: 
\begin{enumerate}
    \item \textbf{Inference-time Steering:} Does the Generator–Verifier architecture improve success rates at test time compared to standard policies?
    \item \textbf{Policy Improvement:} Does training on self-generated, verified data improve the base policy?
    \item \textbf{Data Efficiency:} How does our autonomous data compare to human-expert data?
\end{enumerate}

We ablate the design of verifier in simulation experiments and find the paradigm can improve the policy performance at inference time, agnostic to the design as well. Based on these results, we select the most effective verifier configuration from simulation for deployment in real-world experiments.


\subsection{Experimental Setup}

\paragraph{Tasks}

We evaluate our method on a suite of simulated and real-world robotic manipulation tasks specified by natural language instructions. In simulation, we focus on 4 manipulation tasks from~\cite{walke2023bridgedata}, following the evaluation protocol of Li et al.~\cite{li24simpler}. These tasks provide a controlled benchmark for assessing policy performance and generalization under standardized conditions. We conduct all real-world experiments on the DROID platform~\citep{khazatsky2024droid}.  We design a set of challenging manipulation tasks that stress visual grounding, spatial reasoning, manipulation that requires high-precision, etc. Specifically, we select 2 manipulation tasks in the real world for each policy.



\paragraph{Policies}


We evaluate our framework using different pretrained robot policies.
In simulation, we employ an open-sourced BridgeData policy~\cite{ren2025open}, denoted as $\pi_0$-Bridge, trained on BridgeData~\cite{walke2023bridgedata} following the architecture of $\pi_0$~\cite{blackpi0}.
In real-world experiments, we use various generalist robot policies. Specifically, $\pi_0$-DROID~\cite{blackpi0}, $\pi_{0.5}$-DROID~\cite{pi05} trained on large-scale real-world demonstration data and fine-tuned on~\cite{khazatsky2024droid}.
Unless otherwise specified, all policies are treated as fixed base policies for inference-time steering and are used as action proposers within the generator–verifier framework.




\paragraph{Baselines}
To show the efficiency of our verifier against several baselines for inference time steering and ablate different verifier designs in the simulation experiments. We compare the \MethodName verifier to several key baselines.
\begin{itemize}
    \item \textbf{V-GPS~\cite{nakamoto2024steering}:} A language-conditioned value function learned via offline RL for policy steering.
    \item \textbf{Heuristic:} A verifier that defines action primitives and use stage transitions for verification.
    \item \textbf{VLM+Constraints:} A verifier that scores action samples by verifying if they follow VLM-proposed reference-based waypoints that resolve to pixel coordinates via tracked objects.
\end{itemize}
We also compare to RoboMonkey~\cite{kwok2025robomonkey}, which uses a learned single-step verifier by fitting a Gaussian on the action distribution. While effective for local corrections, we find that the RoboMonkey verifier's optimization of local single-step action regions limits its application to completely new tasks that require global reasoning. For more details on this baseline, please see Appendix~\ref{app:experiments}.

In real-world experiments, we emphasize the comparison between our SOTA verifier design reflected by the simulation results, against baselines:

\begin{itemize}
    \item \textbf{PIVOT~\cite{nasiriany2024pivot}:} An approach that uses proposed action primitives and Iterative Visual Optimization for robotic tasks. This baseline combines a naïve action-primitive proposer with an optimization-based action selector, highlighting the importance of a policy proposer.
    \item \textbf{V-GPS-DROID:} We train a V-GPS model on the DROID~\cite{khazatsky2024droid} dataset and deploy it on the same robot hardware setup. This experiment evaluates the efficiency of our combined policy proposer and visual verifier against a learned value function.
\end{itemize}

\subsection{Inference-Time Steering.}
\label{sec:results-steering}



\begin{figure*}
    \centering
    \includegraphics[width=\textwidth]{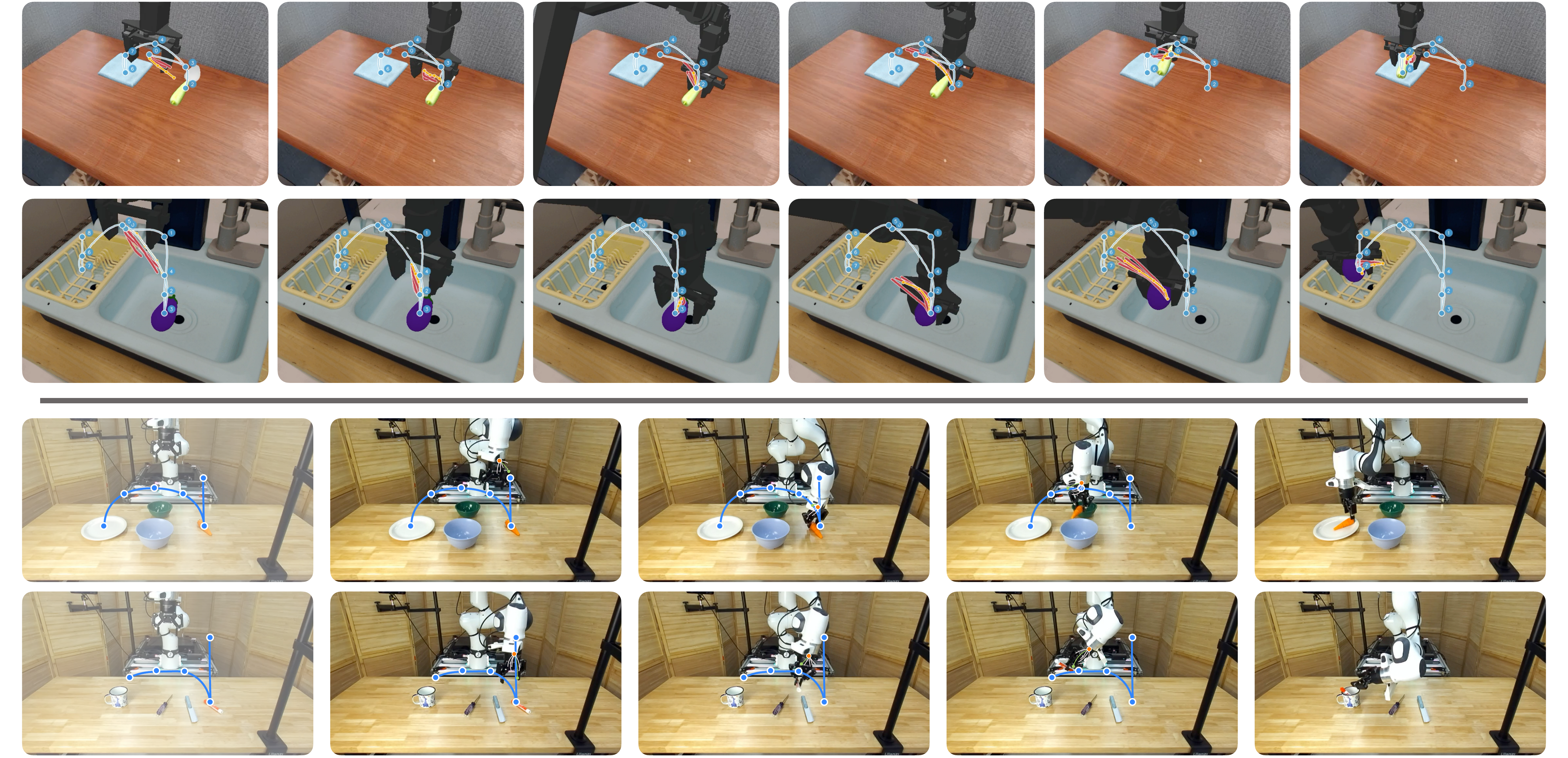}

    \caption{\textbf{Inference-time verification improves task performance.}
    Trajectory rollouts illustrating \MethodName steering. The first 2 rows show verification of the $\pi_0$-Bridge policy on a simulated WidowX and the last 2 rows show verification of the $\pi_0$-DROID policy on a real FR3-DROID robot. At each decision step, the generator generates $N$ candidate action chunks $\{\boldsymbol{a}^{(i)}_{t:t+H}\}_{i=1}^N$; the verifier scores each candidate $V(o_t,\boldsymbol{a}^{(i)}_{t:t+H},l)$ and selects the best action $\boldsymbol{a}^\star$ for execution (highlighted).
    Verifier-guided rollouts achieve successful task completion.}
    \label{fig:steering}
\end{figure*}

In simulation, we evaluate the performance of different verifiers on top of the base policy $\pi_0$-Bridge and take the base policy as one of the baselines in the simulation. All evaluations are averaged over 10 trials on top of task variations such as robot or object locations, etc, following~\cite{li24simpler}. In the real world, we deploy the best-performing verifier in the simulation and evaluate it on 2 different policies using the DROID platform, conducting 50 rollouts for each task across 2 tasks per policy.

\begin{figure*}[ht]
\centering

\begin{minipage}[t]{0.48\textwidth}
  \centering
  \includegraphics[width=\linewidth]{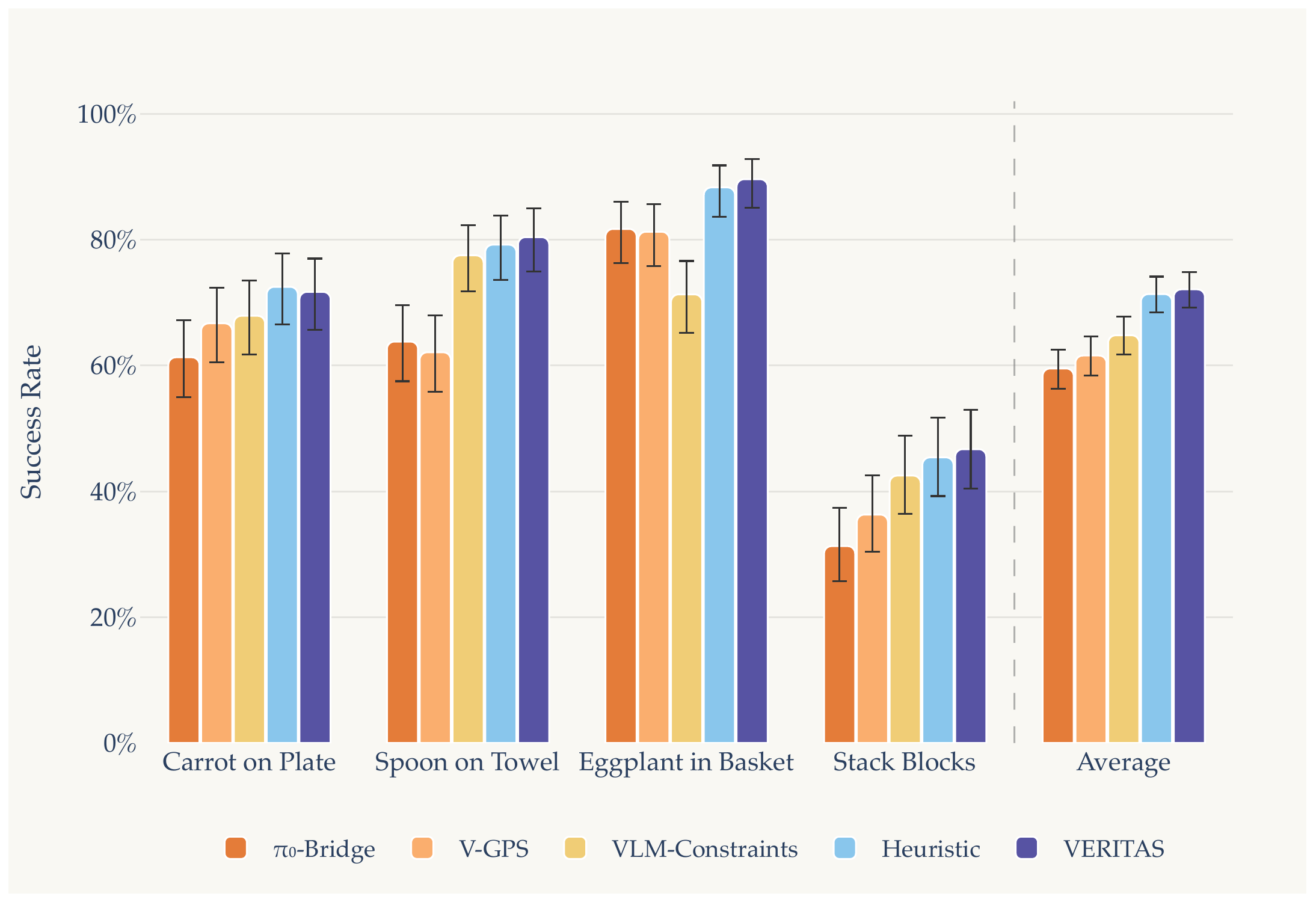}
  \caption{\footnotesize
  \textbf{Inference-time steering in SIMPLER.}
  Simulation success rates comparing standard policy execution and \MethodName execution across 4 tasks and different verifier architectures. \MethodName verifier consistently improves performance over different methods across all tasks. The largest gains are observed in the most challenging task (Stack Blocks). Error bars show 95\% finite-sample CI~\cite{vincent2024generalizable}.
  }
  \label{fig:sim_online}
\end{minipage}
\hfill
    \begin{minipage}[t]{0.48\linewidth}
        \centering
        \includegraphics[width=\linewidth]{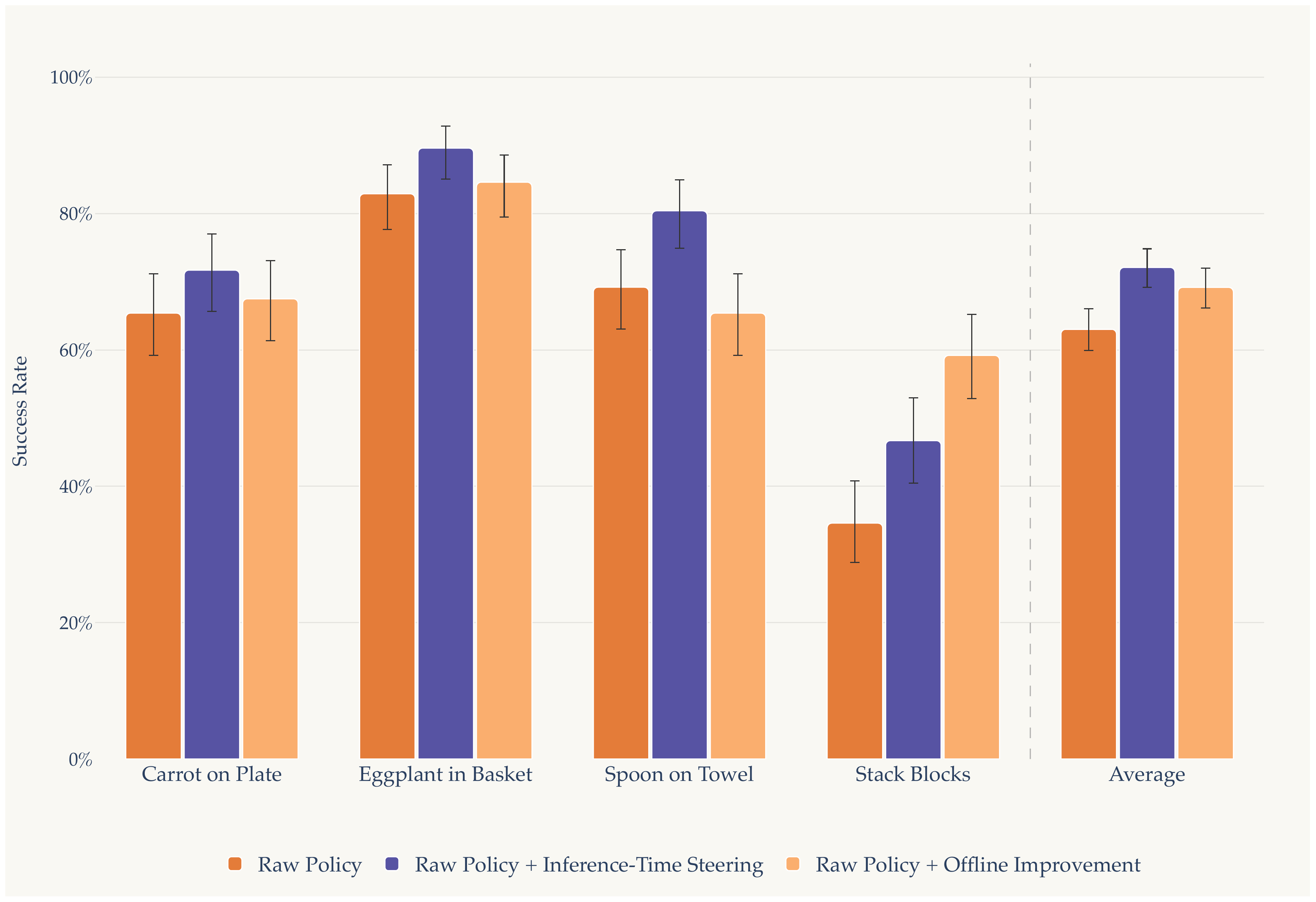}
        \caption{Simulation success rates of the raw policy, the inference-time steering and the same policy fine-tuned on verifier curated autonomous data. While verification improves performance at inference time through action steering, fine-tuning on the collected verifier curated rollouts successfully distills these gains back into the policy weights. Error bars show 95\% finite-sample CI~\cite{vincent2024generalizable}.}
        \label{fig:sim_finetune}
    \end{minipage}
\end{figure*}

As shown in Figure~\ref{fig:sim_online} across all 4 simulation evaluated tasks, verifier-guided action selection consistently improves success rates relative to the base policy. While the magnitude of improvement varies with task difficulty and verifier design, performance gains are observed even when using simple heuristic verifiers, indicating that the benefits of the generator–verifier paradigm are not dependent on a specific verifier architecture. On average, all of our verifier designs outperform V-GPS, suggesting that explicit verification of action consequences provides a more reliable signal for policy steering than the learned value estimates.



\begin{figure*}[t]
    \centering
    \includegraphics[width=\textwidth]{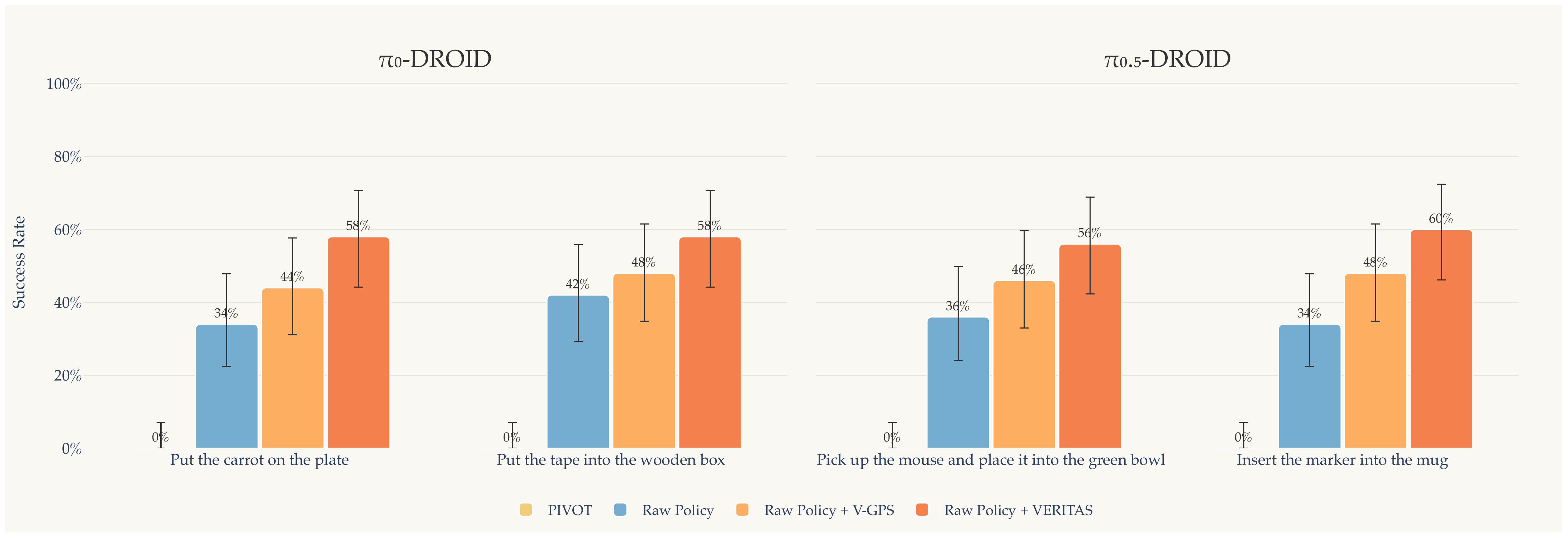}

    \caption{\textbf{Inference-time steering in the real world.} We evaluate inference-time steering across two real-world policies and two tasks per policy, with 50 trials conducted for each task. \MethodName consistently outperforms the V-GPS baseline across all tasks, demonstrating robust improvements in the real world.}
    \label{fig:steer-real}
\end{figure*}

We observe similar trends in real-world experiments. As shown in Figure~\ref{fig:steer-real}, inference-time steering significantly improves task success rates across different evaluated real-world tasks and policies. 

Across simulation and real-world settings, covering 3 different policies and a total of 1160 evaluation episodes, verifier-guided execution improves policy success rates by an average of 12.6\% in simulation and 35\% in real-world deployment, without any policy fine-tuning. These gains highlight the effectiveness of inference-time computation alone in improving generalist policy performance. Note that PIVOT’s naïve action primitives can not achieve successful outcomes, whereas a trained policy with a strong action prior performs effectively, highlighting the importance of good action priors for improving performance.

Importantly, the proposed framework is policy-agnostic and verifier-agnostic. We observe consistent improvements across different base policies and across a diverse set of verifier implementations in simulated and real-world environments. This robustness demonstrates that the generator–verifier framework provides a general and reliable mechanism for inference-time policy steering, applicable across tasks, embodiments, and verification strategies.

\subsection{Offline Policy Improvement}
\label{sec:results-improvement}
We next study whether verified autonomous executions can be used to improve the policy offline.
In this setting, we deploy the policy using the generator--verifier framework in the simulation and real world to collect a set of verified successful trajectories for offline policy fine-tuning.

Specifically, in the simulation experiments, we select the verifier that boosts the policy steering performance the most and collect the successful verified rollouts, consisting of 656 demonstrations for the base policy fine-tuning. As shown in Figure~\ref{fig:sim_finetune}, fine-tuning on this autonomously collected data improves average policy performance and consistently outperforms the base policy across all 4 tasks. 

Notably, inference-time steering provides larger performance gains on tasks that are difficult for the base policy, such as Stack Blocks, than on simpler tasks. One possible explanation is that the policy produces more diverse action samples when it is uncertain or underperforms, creating greater opportunity for the verifier to select better candidates. In contrast, for tasks on which the policy already performs well, the sampled actions are less diverse, leaving less room for improvement. These results point to a fundamental trade-off: the effectiveness of inference-time steering depends on the diversity of the proposed actions, and therefore tends to be greatest on more challenging tasks.


In the simulation experiments, \MethodName fine-tuning improves average success rate by 9.7\% over the base $\pi_0$-Bridge policy across 4 simulated manipulation tasks in SimplerEnv for 960 total episodes. Success rate gains are consistent across all tasks, with the largest improvement on Stack Blocks, from 31.3\% to 59.2\%, by 27.9\% on the hardest task. In the real world experiments shown in Figure~\ref{fig:data_efficiency}, verified autonomous data consistently supports effective policy improvement across 4 challenging real-world tasks for 2 policies, with performance improving reliably as more data is collected, demonstrating that autonomous rollouts alone are sufficient for successful fine-tuning. These results demonstrate that verified autonomous executions can serve as a reliable and effective source of supervision for offline policy improvement, without requiring additional human demonstrations. By leveraging the generator–verifier framework to curate high-quality rollouts, \MethodName consistently improves policy performance.

\subsection{Autonomous v/s Human-Collected Data}
\label{sec:data_efficiency}

Finally, we compare the data efficiency of verifier-curated autonomous rollouts to human expert demonstrations as a supervision signal for offline policy improvement.
While human demonstrations remain a strong baseline, they are expensive to collect and difficult to scale.
An important question is whether autonomously collected, verifier-selected data can provide comparable learning signal.

\begin{figure*}[h]
    \centering
    \includegraphics[width=\textwidth]{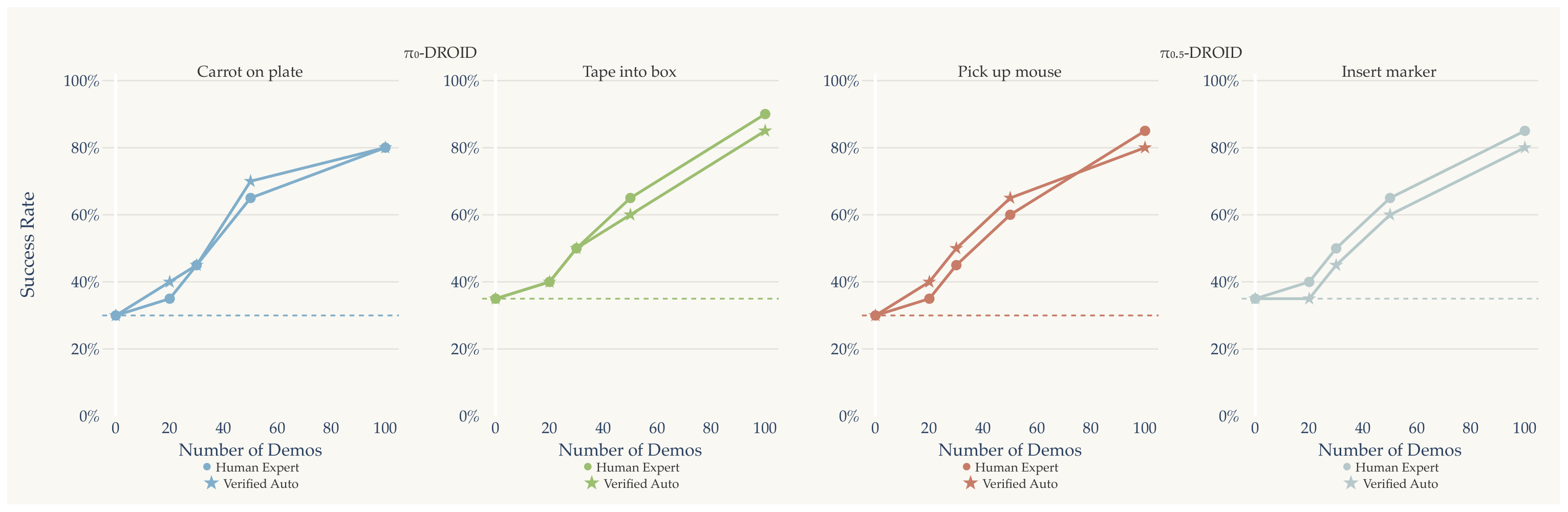}

    \caption{\textbf{Comparing data efficiency of autonomous verification v/s human expert}. We fine-tune the base policy on datasets of 20, 30, 50, and 100 trajectories collected via either human expert teleoperation or verifier-curated autonomous rollouts, and evaluate each resulting policy over 20 real-world trials. Across all four tasks, verifier-curated data achieves comparable or superior data efficiency to human demonstrations. In multiple cases, autonomous data matches or outperforms human data at the same budget (e.g., 20--50 demos), and remains competitive even at larger scales. For example, on \emph{Pick up mouse}, 50 autonomous demos outperform human demonstrations (0.65 vs. 0.60), while for \emph{Carrot on plate}, performance is identical. These results indicate that verifier-curated autonomous collection can match the
   data efficiency of expert demonstrations, enabling scalable policy improvement without supervision.}
    \label{fig:data_efficiency}
    \vspace{-10pt}
\end{figure*}

In our real-world experiments, we fine-tune policies using datasets of 20, 30, 50, and 100 demonstrations collected from either verifier-curated autonomous rollouts or human expert teleoperation. We then evaluate each fine-tuned policy over 20 real-world trials across multiple manipulation tasks, comparing policy performance as a function of the quantity and source of fine-tuning data. This controlled setup allows us to directly compare how policy performance scales with the amount and source of fine-tuning data.

The experimental results are shown in Figure~\ref{fig:data_efficiency}. Across all 4 tasks, with 2 tasks evaluated for each policy, verifier curated data demonstrates comparable data efficiency to human expert demonstrations. In several cases, autonomous data achieves stronger performance at moderate data budgets. For example, on \textit{Carrot on plate}, fine-tuning with 50 autonomous demonstrations outperforms fine-tuning with the same number of human demonstrations (0.70 vs.~0.65). Similarly, on \textit{Pick up mouse}, autonomous data consistently achieves higher success rates in the 20 to 50 demonstration regime.

At larger data budgets, verifier curated data remains competitive with expert supervision across all tasks. On \textit{Carrot on plate}, the two data sources converge to the same success rate at 100 demonstrations, while on \textit{Insert marker}, the performance gap remains within 5 percentage points across all data budgets. Overall, these results show that verifier curated autonomous rollouts provide an effective and scalable alternative to costly human demonstrations for policy improvement.


Overall, these results show that verifier-curated autonomous rollouts can match the data efficiency of costly human teleoperation.
By converting deployment-time execution into effective training data, our approach enables scalable policy improvement without requiring continuous expert supervision. These findings suggest that execution-time verification not only improves performance online, but also enables a practical pathway for continual policy improvement by converting deployment experience into effective training data.

More importantly, because the robot itself generates the data, the autonomously collected dataset is inherently on-policy and kinematically feasible. The verifier serves as an expert filter, retaining high-quality executions while discarding failures and suboptimal behaviors. This process effectively closes the learning loop: the reasoning and corrective signals provided by the verifier are distilled back into the policy during offline fine-tuning, progressively improving the generator and reducing the need to sample a large number of candidates in future deployments.

\subsection{Implementation Hyperparameters}
\label{exp:ablation}

\begin{figure*}[t]
    \centering

    \begin{minipage}[t]{0.49\linewidth}
        \centering
        \includegraphics[width=\linewidth]{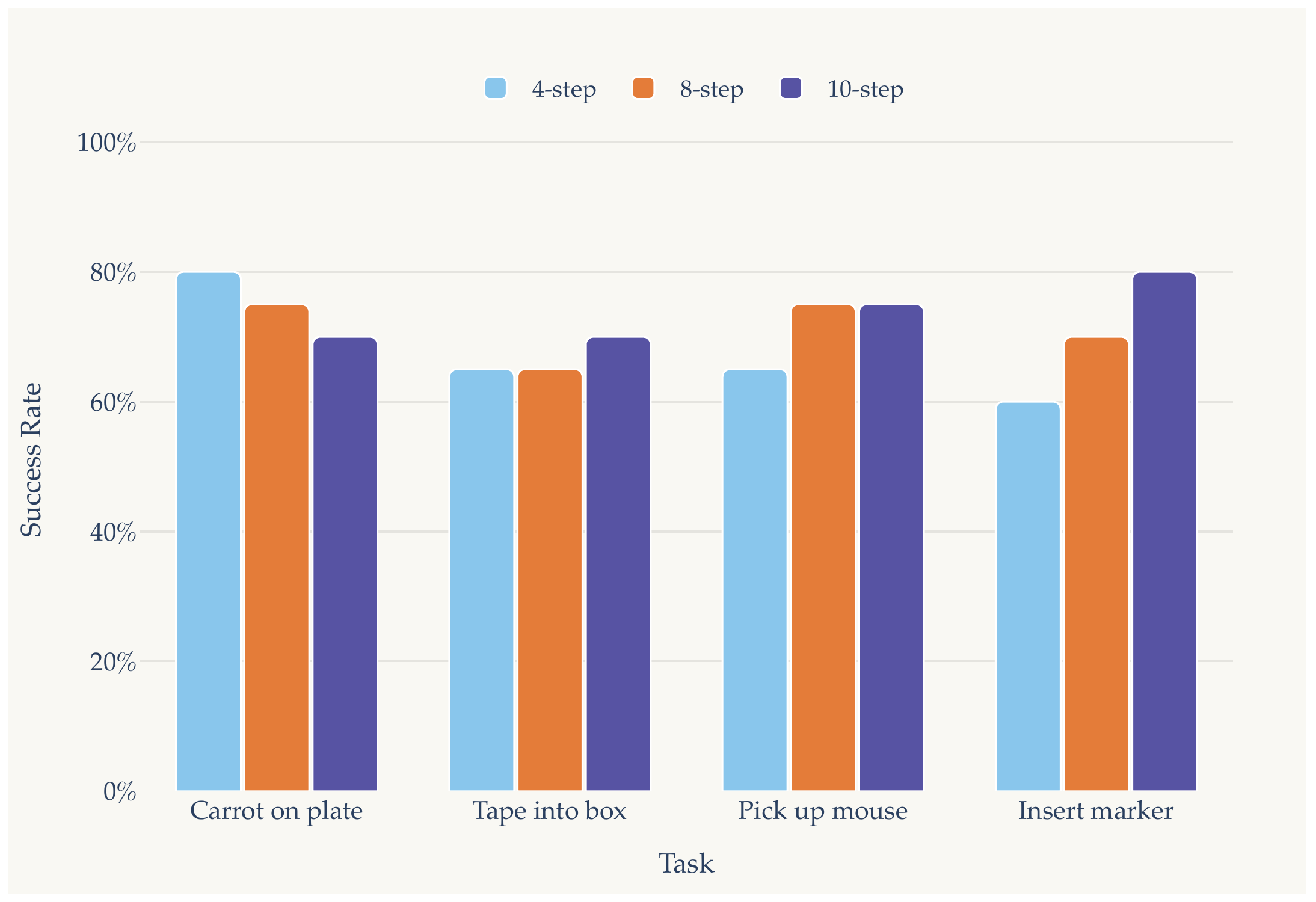}
        \caption{\footnotesize
        We find that VERITAS performance does not significantly degrade with open-loop action execution of up to 10 steps.
        }
        \label{fig:step-ablation}
    \end{minipage}
    \hfill
    \begin{minipage}[t]{0.49\linewidth}
        \centering
        \includegraphics[width=\linewidth]{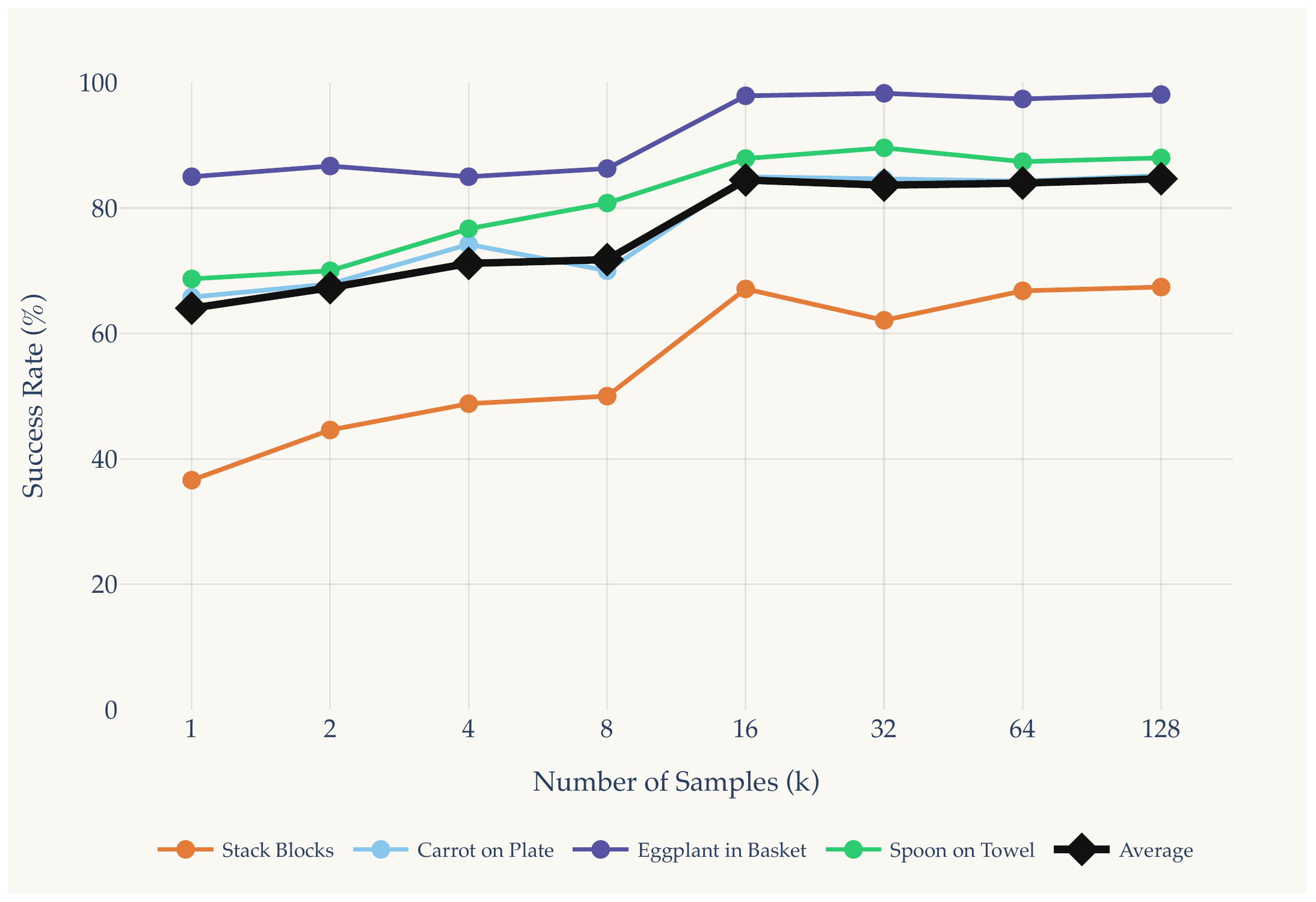}
        \caption{\footnotesize
        We find that inference-time verification performance saturates beyond $N=8$ samples, while the cost of sampling increases linearly with $N$.
        }
        \label{fig:sample-abation}
    \end{minipage}
\end{figure*}



We perform a sweep over the action execution horizon, which determines the length of the visual action trace associated with each action chunk (Figure~\ref{fig:step-ablation}). For $\pi_0$-DROID, we find that the open-loop execution horizon does not significantly degrade verification performance in our real-world experiments. Based on this observation, we use the default horizon of 8 steps for all real-world evaluations.

We also sweep over the number of action samples $N$ on task success rate in simulation. As shown in Figure~\ref{fig:sample-abation}, steering performance generally improves as the number of action samples increases, suggesting that larger candidate sets provide better opportunities for verification. However, we find that performance saturates beyond $N=8$, serving as a practical guide for real-world deployment.


\section{Discussion}

In this work, we introduced a generator-verifier framework that transforms inference-time computation into sustainable policy improvement. By decoupling reasoning from action generation, we demonstrated that pre-trained generalist policies can achieve substantial performance gains purely through inference-time steering. We further showed that successfully verified trajectories can be distilled into the base policy via post-training to improve the policy performance in a data-efficient manner. Across a range of simulated and real-world benchmarks, we demonstrate consistent performance improvements with the VERITAS with 20--100 trajectories of experience. Most notably, we find that autonomous verification matches the data efficiency of human-collected expert trajectories. This creates a scalable pathway for robot self-improvement, where deployment experience naturally translates into persistent capability improvements.

While effective, our framework relies on trading inference compute for task performance, utilizing repeated sampling that may be computationally expensive in latency-critical applications. Additionally, our current verifier implementation depends on static visual traces generated at the onset of the task; while this is sufficient for quasi-static manipulation, it may struggle in highly dynamic environments where the scene changes rapidly during execution. Future work could address these constraints by investigating mechanisms to distill the verifier’s rejection logic into a value function for faster inference, or by exploring joint optimization of the generator and verifier to improve sample efficiency during the search process. Further, it is important to note that our verifier can only improve performance by selecting the best action candidate among those proposed by the policy. This is fundamentally limited by the pre-trained exploration prior learned by the policy, and we expect improvements in policy pre-training to further improve the verification pipeline.

\section*{Acknowledgments}
This research was partially supported by the Toyota Research Institute and Microsoft Research, with compute support from Google TPU Research Cloud, NVIDIA Academic Grant Program, and Gemini Academic Program. The authors also thank Mitsuhiko Nakamoto, for help with reproducing V-GPS, and Samuel M. Bateman, Yanbo Xu, Hengkai Pan for helpful discussions.

\clearpage
\bibliographystyle{unsrt}
\bibliography{references.bib}

\clearpage
\beginappendix{
\section{Implementation Details}
\subsection{Visual Verifier Design}
\label{app:verifier}


We mainly design and ablate 3 types of verifier design in our paper. We discuss the design of each of them:

We denote the RGB observation at time $t$ as $F_t$, and the end-effector pixel location as $e_t \in \mathbb{R}^2$. 
Let $\mathcal{I}$ denote the natural-language task instruction.

Object detections obtained from detection~\cite{comanici2025gemini, ravi2025sam} or tracking models~\cite{ren2024grounded} are denoted as $\hat{o}_t$, where each detection contains a bounding box and a center pixel coordinate. 
A set of tracked objects is denoted by $\mathcal{O}$.

A waypoint sequence is denoted by $\mathcal{W} = \{w_1, \dots, w_K\}$, where each waypoint may be expressed either as:
\begin{itemize}
\item An absolute pixel coordinate $w_i = (u_i, v_i)$, or
\item A reference expression (e.g., \texttt{role:source}, \texttt{midpoint(source,target)}) that resolves to a pixel location.
\end{itemize}

We denote the current active waypoint index as $i$, and the total number of waypoints as $K$. 
The distance between the end-effector and the current waypoint is written as
\[
d_t = \| e_t - g_i \|,
\]
where $g_i$ is the resolved pixel goal for waypoint $i$.

Verification scores are normalized to the range $[0,1]$ and denoted as $s_t$. 
Progress through a waypoint sequence is defined as
\[
p_t = \frac{i}{K}.
\]

For the heuristic verifier, the current stage of the finite-state machine is denoted by $\sigma_t$, and task progress is denoted by $p_t$, computed from geometric features such as distances and object motion.

Hyperparameters include tolerance thresholds $\text{tol}_i$, smoothing windows, and weighting coefficients $\alpha$ and $\tau$ used in score computation.

\paragraph{VERITAS} The assumption for VERITAS design is that: General manipulation tasks can be naturally described as a spatial motion trace of the robot end-effector or gripper: such as approaching an object, moving along a surface, or transporting an object to a target location. While such traces are difficult to specify analytically for diverse tasks, vision-language models (VLMs) are highly capable of imagining plausible execution trajectories directly from task instructions and images.

The VERITAS Verifier leverages this capability by asking a VLM to generate a single, absolute-pixel waypoint trace at the beginning of an episode. During execution, the verifier checks whether the robot’s end-effector/gripper follows this plan with sufficient fidelity.

Specifically, the verifier follows a simple principle: If the robot’s end-effector motion stays close to a plausible pixel-space trajectory imagined by a VLM that has strong embodied reasoning capability, the behavior is likely correct to complete the manipulation tasks. This design avoids explicit task stages, object-specific heuristics, or per-frame reasoning. Instead, it evaluates execution by measuring geometric consistency between the observed object trajectory and a generated waypoint visual trace.

\begin{itemize}
    \item \textbf{Initialization: VLM-Generated Waypoint Trace.} Given the task instruction and the initial image observation, the VLM identifies which objects in the scene are relevant to the tasks and assigns them semantic roles (e.g., source, target). These objects are used \textit{only to provide spatial hints} through object detection and to help the VLM ground its reasoning. Second, using the same instruction and image with object detection results provided, the VLM generates a sequence of absolute pixel-space waypoints that describe how the robot should move to accomplish the task. Each waypoint specifies: a target pixel location $(u, v)$, a tolerance radius in pixels along with optional constraints such as minimum dwell time or whether the waypoint may be skipped. 
    \item \textbf{Execution-Time Verification.} During execution, the verifier observes the robot’s end-effector or gripper pixel location at each timestep. Verification is performed by checking whether the EEF/gripper: reaches each waypoint within its tolerance in the image space, respects ordering constraints between waypoints, maintains proximity for required durations when specified.
\end{itemize}

As described in Algorithm~\ref{alg:waypoint_px_verifier}, the verifier does not encode task-specific logic or any predefined action primitives (e.g., grasp, place, open). All task knowledge is captured implicitly in the waypoint trace generated by the VLM. It is robust to partial execution as these waypoints can be marked as skippable or constraints are soft, allowing the verifier to tolerate minor deviations, alternative but valid execution styles.

\begin{algorithm}
\caption{\MethodName Verifier}
\label{alg:waypoint_px_verifier}

\KwIn{Instruction $\mathcal{I}$, initial frame $F_0$, EEF pixels $\{e_t\}$}
\KwOut{Verification score $s_t \in [0,1]$}

$\mathcal{W} \leftarrow \text{VLM Waypoint Trace}(\mathcal{I}, F_0)$\;
Clamp $|\mathcal{W}| \in [5,10]$, enforce start/end non-skippable\;
$i \leftarrow 0$\;

\For{$t = 1 \ldots T$}{
    $d_t \leftarrow \| e_t - \mathcal{W}[i].uv \|$\;
    
    \If{$d_t < \text{tol}_i$ for $\geq$ min\_hold frames}{
        $i \leftarrow i + 1$\;
    }
    
    $p_t \leftarrow i / |\mathcal{W}|$\;
    $s_t \leftarrow \alpha p_t + (1-\alpha)\exp(-d_t / \tau)$\;
}

\If{$i \geq |\mathcal{W}|$}{
    $s_t \leftarrow 1.0$\;
}
\end{algorithm}

\paragraph{VLM-Constraints Verifier}

Similarly, we have a VLM-Constraints Verifier design that similar to VERITAS but with dynamic reference resolution. It evaluates robot behavior by comparing the executed robot motion against a reference execution trace generated by a vision-language model (VLM). Unlike absolute pixel-space waypoints, the reference trace is expressed as a sequence of relative, object-centric waypoint references that are dynamically resolved using object tracking during execution. The key insight is to separate what the robot should do from how the robot actually moves, and to verify correctness by checking whether execution follows a plausible reference trace. For general manipulation tasks, correctness is best judged not by low-level actions or final success alone, but by whether the robot follows a reasonable sequence of spatial intentions: approach the source object, move toward a target, align with an intermediate region, and so on.

\begin{itemize}
    \item \textbf{Reference-Based Waypoints Initialization.} Similar to VERITAS, at the start of an episode, a VLM is queried with the task instruction and the initial observation. Note that instead of returning pixel coordinates, the VLM produces a list of objects to track, each with a semantic role (e.g., source, target), and a sequence of waypoints defined by references. These waypoint references specify what the robot should aim for, not where that is in pixels. This abstraction allows the reference trace to remain valid even as objects move during execution.
    \item \textbf{Dynamic Resolution via Tracking.} During execution, the verifier runs an object tracker continuously. At every frame, each waypoint reference is resolved into a pixel-space goal using the latest tracked object positions. The composite references (e.g., midpoints) are resolved by combining multiple tracked entities and he resulting waypoint locations move with the scene, automatically adapting to object motion, occlusion, or partial interaction.
    \item \textbf{Execution-Time Verification.} The verifier maintains an index over the waypoint sequence and checks, at each frame, whether the EEF is sufficiently close to the current waypoint. Progression through waypoints is governed by simple but robust mechanisms such as distance thresholds normalized by object scale, hysteresis to avoid oscillation near boundaries and minimum hold time to ensure stable attainment.
\end{itemize}

Once all waypoints are completed, the task is considered done. Importantly, the verifier does not enforce a single rigid execution style; it tolerates variations as long as the robot follows the intended reference trace. The verifier explicitly encodes a reference answer, which is a plausible execution trace generated from VLM and checks whether the robot’s behavior aligns with it.

\begin{algorithm}
\caption{VLM-Constraints Verifier}
\label{alg:tracking_waypoint_verifier}

\KwIn{Instruction $\mathcal{I}$, frames $\{F_t\}$, EEF pixels $\{e_t\}$}
\KwOut{Verification score $s_t \in [0,1]$}

$(\mathcal{O}, \mathcal{W}) \leftarrow \text{VLM Waypoint Trace}(\mathcal{I}, F_0)$\;
Initialize tracker with objects $\mathcal{O}$\;
Clamp $|\mathcal{W}| \in [5,10]$\;
$i \leftarrow 0$\;

\For{$t = 1 \ldots T$}{
    $\hat{o}_t \leftarrow \text{Tracker}(F_t)$\;
    $g_i \leftarrow \text{ResolveRef}(\mathcal{W}[i].ref, \hat{o}_t)$\;
    $d_t \leftarrow \| e_t - g_i \| / \text{scale}(g_i)$\;
    
    \If{$d_t < \text{tol}_i$ for $\geq$ min\_hold frames}{
        $i \leftarrow i + 1$\;
    }
    \ElseIf{stalled \textbf{and} next waypoint skippable}{
        $i \leftarrow i + 1$\;
    }
    
    $p_t \leftarrow i / |\mathcal{W}|$\;
    $s_t \leftarrow \alpha p_t + (1-\alpha)\exp(-d_t / \tau)$\;
}

\If{$i \geq |\mathcal{W}|$}{
    $s_t \leftarrow 1.0$\;
}
\end{algorithm}

\paragraph{Heuristic Verifier}

The motivation for the Heuristic Verifier design is that, we assume most manipulation tasks can be decomposed into a small number of stages (approach, align, engage, manipulate, release, etc.) and that correct behavior looks like progressing through these stages in order.

Specifically at its core, the verifier models execution as progression through a sequence of stages. We formalize this assumption using a finite-state machine (FSM) that tracks the current stage of the task based on simple geometric and kinematic signals:

\begin{align*}
    \texttt{APPROACH} \rightarrow \texttt{ALIGN} \rightarrow \texttt{ENGAGE} \rightarrow \texttt{MANIPULATE}
\rightarrow \texttt{RELEASE} \rightarrow \texttt{SETTLE} \rightarrow \texttt{DONE}.
\end{align*}

Each stage corresponds to a qualitatively different interaction regime. For example, \texttt{APPROACH} expects the end-effector to move closer to a relevant object, \texttt{ENGAGE} expects grasping or contact behavior, and \texttt{MANIPULATE} expects the object to move toward a target. The verifier does not prescribe how these stages are achieved; it only checks whether the observed behavior is consistent with being in the current stage and whether transitions occur in a plausible order.

During execution, the verifier continuously tracks the robot end-effector and relevant objects in the image. From these observations, it computes lightweight geometric features such as end-effector–object distances, object–object distances, relative motion between objects and the end-effector, and gripper open/close state. These features are temporally smoothed to reduce noise and serve as input to the stage transition logic. A finite-state machine consumes the extracted features and determines (1) which stage the robot is currently in, and (2) whether it is appropriate to remain in that stage or transition to the next one. Transitions are triggered by interpretable conditions such as the end-effector being sufficiently close to an object, an object moving consistently with the end-effector (indicating grasp), or the manipulated object approaching a target location. This structure enforces a coarse notion of temporal correctness without requiring explicit trajectory specification. In addition to discrete stage transitions, the verifier computes a continuous progress signal based on task-specific geometric measures, such as decreasing distance to a target. Changes in this signal are tracked over time to detect forward progress, stagnation, or regression. If progress stalls for a sustained period, the verifier can optionally invoke a vision-language model to diagnose potential failure modes (e.g., wrong object, misalignment, or obstruction) and suggest recovery actions. Please see~\ref{alg:fsm} for more details and pesudo code of the finite state machine.

\begin{algorithm}
\caption{Stage FSM Update}
\label{alg:fsm}
\KwIn{$s$, engaged, released, progress\_good, is\_grasped, task\_done}
\KwOut{$s$}

\eIf{$s=\textsc{Approach}$}{
    \If{$engaged \lor is\_grasped$}{
        $s \gets \textsc{Align}$\;
    }
}{
\eIf{$s=\textsc{Align}$}{
    \If{$is\_grasped \land progress\_good$}{
        $s \gets \textsc{Manipulate}$\;
    }
    \ElseIf{$engaged \land progress\_good$}{
        $s \gets \textsc{Engage}$\;
    }
    \ElseIf{$\neg engaged$}{
        $s \gets \textsc{Approach}$\;
    }
}{
\eIf{$s=\textsc{Engage}$}{
    \If{$progress\_good$}{
        $s \gets \textsc{Manipulate}$\;
    }
    \ElseIf{$\neg engaged \land \neg is\_grasped$}{
        $s \gets \textsc{Align}$\;
    }
}{
\eIf{$s=\textsc{Manipulate}$}{
    \If{$released$}{
        $s \gets \textsc{Release}$\;
    }
}{
\eIf{$s=\textsc{Release}$}{
    \If{$progress\_good$}{
        $s \gets \textsc{Settle}$\;
    }
    \ElseIf{$\neg released$}{
        $s \gets \textsc{Manipulate}$\;
    }
}{
\If{$s=\textsc{Settle} \land task\_done$}{
    $s \gets \textsc{Done}$\;
}
}}}}}

\Return{$s$}
\end{algorithm}

\begin{algorithm}
\caption{Heuristic Verifier}
\label{alg:heuristic_verifier}

\KwIn{Instruction $\mathcal{I}$, frames $\{F_t\}$, EEF pixels $\{e_t\}$, gripper states $\{g_t\}$}
\KwOut{Verification score $s_t \in [0,1]$, stage $\sigma_t$}

$(\text{task\_type}, \mathcal{O}) \leftarrow \text{VLM Router}(\mathcal{I}, F_0)$\;
Initialize tracker with objects $\mathcal{O}$\;
Initialize FSM stage $\sigma \leftarrow \texttt{APPROACH}$\;

\For{$t = 1 \ldots T$}{
    $\hat{o}_t \leftarrow \text{Tracker}(F_t)$\;
    $\phi_t \leftarrow \text{Extract Features}(e_t, \hat{o}_t, g_t)$\;
    $\sigma_t \leftarrow \text{FSM Update}(\sigma_{t-1}, \phi_t)$\;
    $p_t \leftarrow \text{ComputeProgress}(\text{task\_type}, \sigma_t, \phi_t)$\;
    
    \If{progress stalls}{
        Optional: $\text{VLM Diagnosis}(\mathcal{I}, \{F_{t-k:t}\})$\;
    }
    
    $s_t \leftarrow \text{Score}(\sigma_t, p_t, \phi_t)$\;
}
\end{algorithm}

\subsection{Real-world Setup}

In the real-world experiment setup, policy observations consist of images captured from a wrist-mounted camera and an additional external third-person view camera, following~\cite{khazatsky2024droid}. To enable reliable visual verification, we further deploy a calibrated front-facing camera that observes both the robot arm and the workspace. This viewpoint reduces occlusions and provides a stable perspective for evaluating pixel-space visual traces.

In simulation, the open-loop execution horizon of the $\pi_0$-Bridge policy is set to 4 steps, whereas in real-world experiments, the open-loop execution horizon of the $\pi_0$-DROID policy is set to 8 steps.

Tables~\ref{tab:hyperparams} and~\ref{tab:pi0fast_hyperparams} summarize the key hyperparameters used to fine-tune the $\pi_0$-Bridge and $\pi_0$-DROID policies, respectively. Unless otherwise specified, all remaining training settings follow the default configurations of the original implementations.

\begin{table}[h]
\centering
\small
\caption{$\pi_0$-Bridge finetuning Hyperparameters.}
\label{tab:hyperparams}
\renewcommand{\arraystretch}{1.15}
\begin{tabular}{l c}
\toprule
\textbf{Hyperparameter} & \textbf{Value} \\
\midrule
Global batch size & 1024 \\
Per-device batch size & 16 \\
Learning rate & $5 \times 10^{-5}$ \\
Weight decay & 0 \\
Max grad norm & 1.0 \\
\bottomrule
\end{tabular}
\end{table}

\begin{table}[h]
\centering
\small
\caption{$\pi_0$-DROID Fituning Hyperparameters.}
\label{tab:pi0fast_hyperparams}
\renewcommand{\arraystretch}{1.15}
\begin{tabular}{l c}
\toprule
\textbf{Hyperparameter} & \textbf{Value} \\
\midrule
Warmup steps & 1000 \\
LR decay steps & 1{,}000{,}000 \\
Learning rate & $5 \times 10^{-5}$ \\
Batch size & 256 \\
\bottomrule
\end{tabular}
\end{table}

We provide the prompts for the verifier to generate the visual trace. Specifically, prompts used for visual trace generation for VERITAS:

\begin{tcolorbox}
\begin{verbatim}
Generate 5-10 absolute pixel waypoints (u,v) for the robot end-effector to 
complete the task.
Return JSON matching schema:
waypoints: [{{"uv": [u,v], "tol_px": float, "min_hold": int, 
"skippable": bool, "weight": float}}],
confidence: float, notes: optional.
Instruction: {instruction if instruction else "No instruction provided"}.
pixel coords are absolute.
\end{verbatim}
\label{ref: code_VERITAS}
\end{tcolorbox}

Prompts used for visual trace generation for the VLM-Constraints verifier: 
\begin{tcolorbox}
\begin{verbatim}
Generate a waypoint plan for a robot end effector.
Use ONLY reference-based waypoints 
(ref strings): role:source, role:target, object:<name>, midpoint(refA,refB).

DO NOT return any pixel coordinates. 
Return 5 to 10 waypoints, sequential, each with tol_norm and min_hold 
(frames to hold inside tolerance).
The waypoints should guide approach, alignment, 
and placement using the referenced objects.
\end{verbatim}
\label{ref: code_vlm_base}
\end{tcolorbox}

\section{Additional Experiments}

\begin{figure}
    \centering
    \includegraphics[width=\linewidth]{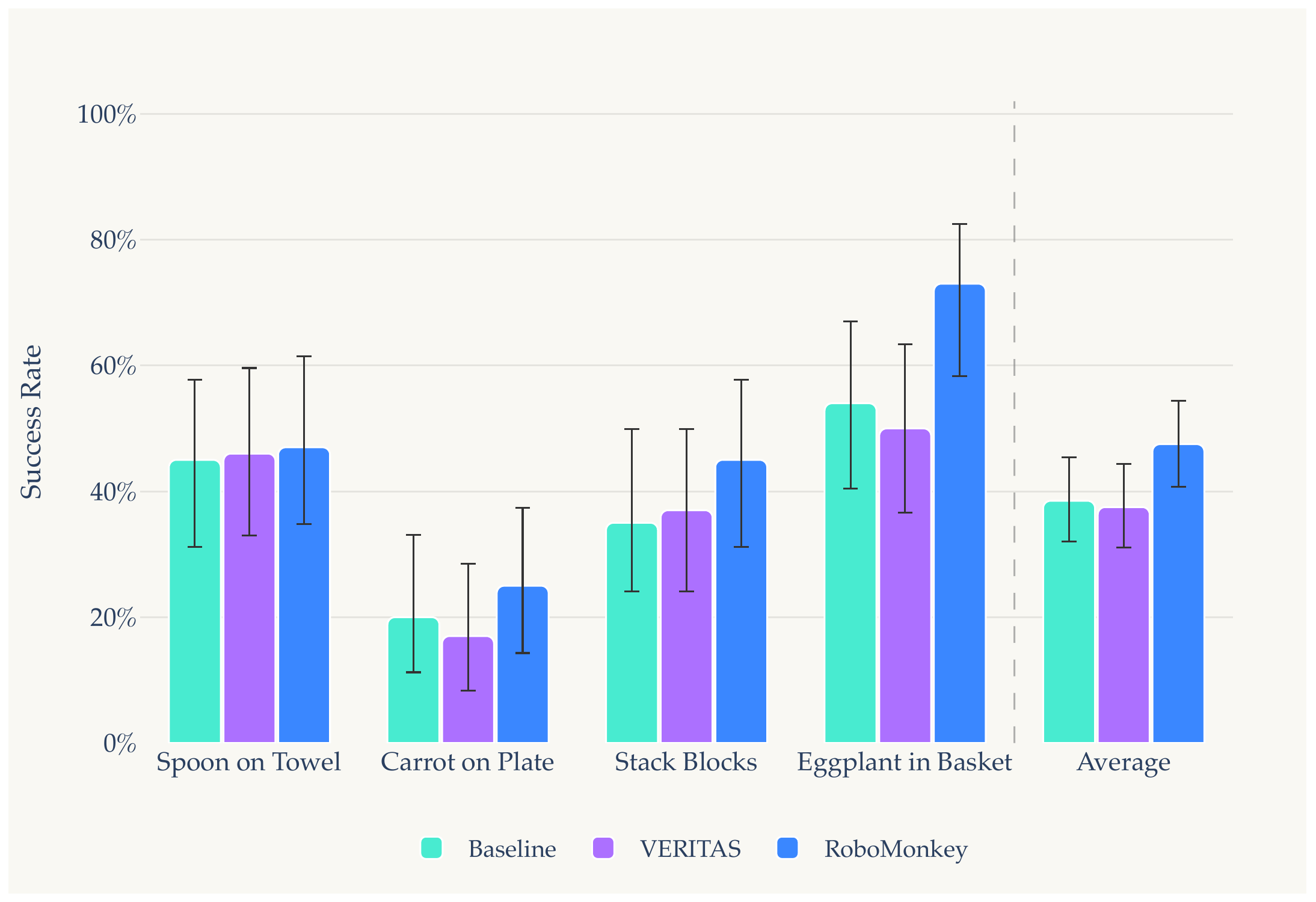}
    \caption{\textbf{Comparision with RoboMonkey.} Simulation success rates for different data augmentation methods across the same 4 simulation tasks (50 episodes per task; RoboMonkey averaged over 3 random seeds). RoboMonkey consistently improves performance over the baseline, achieving an average success rate gains observed on all tasks. In contrast, VERITAS yields mixed outcomes, improving performance on some tasks while degrading others. This suggests that naive data augmentation without enforcing semantic consistency across action chunks can negatively impact performance. Note that this comparison uses single-step action perturbations rather than the action-chunk-level steering used in Figure 4; the aggregate performance difference is dominated by architectural mismatch rather than the algorithm.}
    \label{fig:exp_robomonkey}
\end{figure}



\begin{figure*}[t]
    \centering
    \includegraphics[width=0.95\linewidth]{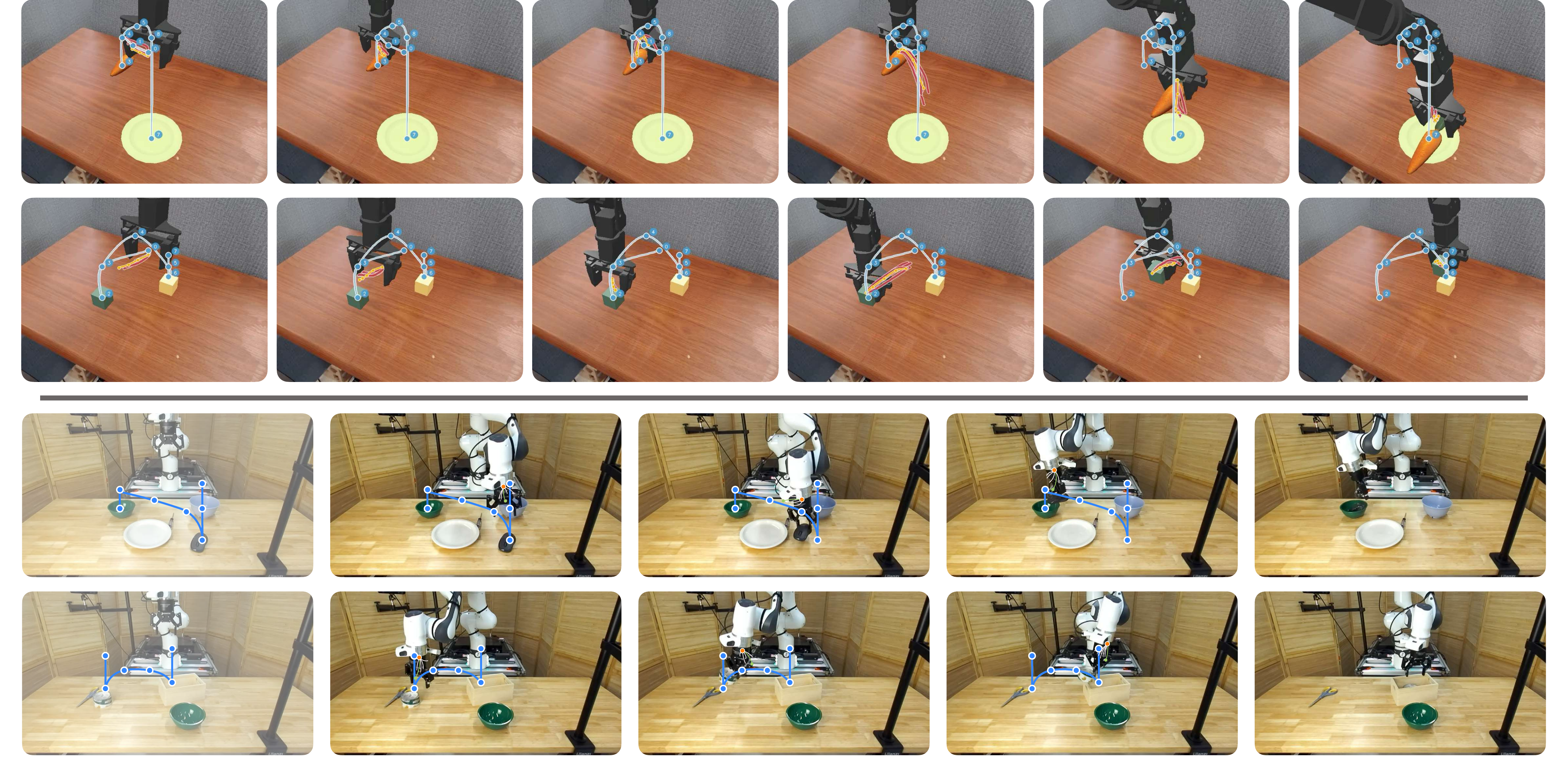}
    \caption{\footnotesize
    Additional results of inference-time verification improves task performance.
    }
    \label{fig:more quali}
\end{figure*}




\label{app:experiments}
\paragraph{Additional Experiments on RoboMonkey}

As discussed in our experiments, we also compare VERITAS with RoboMonkey for online policy steering, with results shown in Figure~\ref{fig:exp_robomonkey}. RoboMonkey demonstrates that, across a range of VLAs, the relationship between action error and the number of generated samples follows an exponentiated power law. Its verifier quantifies the discrepancy between VLA-generated actions and actions from an offline dataset, effectively measuring how closely a sampled action matches previously observed policy behavior.

However, RoboMonkey operates at the level of single-step action perturbations, typically by applying Gaussian noise around a policy output. This form of exploration does not capture high-level action reasoning or alternative strategies. Instead, the verifier optimizes a local ridge in the action space, refining small deviations of a single behavior rather than comparing distinct reasoning traces. Because the verifier is trained on policy-like actions, it implicitly learns and reinforces the policy’s existing biases, and does not distinguish between qualitatively different strategies for accomplishing a task.

In contrast, our verifier evaluates entire action chunks sampled at a higher temporal level, enabling comparison across diverse action sequences that correspond to different high-level plans. Rather than improving micro-scale action noise, our approach selects among semantically distinct execution traces, allowing the system to reason over alternative strategies instead of merely refining low-level control, as demonstrated in our inference-time steering results such as~\ref{fig:sim_online} and~\ref{fig:steer-real}.

\paragraph{More qualitative results}
We show more quantitative results in Figure~\ref{fig:more quali}, to demonstrate the procedure of inference-time steering by verification in detail.

}

\end{document}